\documentclass[sigconf]{acmart}
\usepackage{fvextra}
\setcopyright{none}
\renewcommand\footnotetextcopyrightpermission[1]{}

\AtBeginDocument{%
  }

\copyrightyear{2026}
\acmYear{2026}
\acmDOI{}
\acmISBN{}

\usepackage{listings}

\begin{document}

\title{Discovery Agents for Real-Time Analytics:\\Toward Proactive Insight Systems}

\author{Gaetano Rossiello}
\affiliation{%
  \institution{IBM}
  \city{New York City}
  \state{New York}
  \country{USA}
}

\author{Dharmashankar Subramanian}
\affiliation{%
  \institution{IBM}
  \city{Yorktown Heights}
  \state{New York}
  \country{USA}
}

\renewcommand{\shortauthors}{Rossiello et al.}

\begin{abstract}
Modern analytics systems are fundamentally reactive, requiring users to define queries over increasingly complex and continuously evolving data. In real-time streaming environments, this paradigm breaks down, as the space of potential insights becomes too large to enumerate manually.
We present a multi-agent architecture for autonomous insight discovery over real-time data streams. The system implements a continuous discovery loop in which agents generate hypotheses, compile them into executable analytics, validate generated artifacts, and produce visualizations and deployable applications. The architecture leverages Apache Kafka for event-driven coordination, Apache Flink for stream processing, and large language models to implement specialized agents.
A key contribution is a contract-driven design based on typed intermediate artifacts, enabling modularity, observability, lineage, and safer execution of dynamically generated analytics. Through use cases in retail, finance, and public data, we show how this architecture supports a shift from query-driven analytics to proactive, discovery-driven systems.

\end{abstract}

\maketitle

\section{Introduction}

Data analytics systems have evolved from manual SQL workflows to dashboards and, more recently, LLM-powered copilots. Despite these advances, the dominant interaction model remains largely unchanged: analytics is still \emph{query-driven}. Users formulate questions, and systems translate them into computations, visualizations, or reports.
This model assumes that users know what to ask. In modern data environments, this assumption increasingly fails. Organizations operate over high-dimensional, heterogeneous, and continuously updated data sources. In particular, real-time streaming systems such as Apache Kafka~\cite{kreps2011kafka,garg2013apache} introduce a dynamic setting in which new patterns may emerge continuously, while the space of possible analytical questions grows combinatorially. As a result, many insights remain undiscovered not because they are difficult to compute, but because no analyst explicitly formulated the corresponding hypothesis.

Recent work on LLM-powered data agents has begun to automate parts of the analytics lifecycle, including data preparation, code generation, visualization, insight generation, and reporting~\cite{rahman2025datascienceagents,sun2025llmstatistics,zhu2025dataagents,fu2025autonomousdataagents}. Existing systems demonstrate the potential of agents for business intelligence, dashboard generation, insight management, heterogeneous data analysis, and multi-agent reasoning~\cite{zhang2025datatodashboard,jiang2025siriusbi,weng2025insightlens,sun2025agenticdata,abaskohi2025agentada,liu2025datasage}. Benchmarks for data-driven discovery and analytics agents further show that multi-step insight generation, hypothesis formulation, and robust execution are becoming central evaluation problems~\cite{majumder2025discoverybench,gu2024blade,sahu2024insightbench}. However, most existing work primarily targets batch datasets, user-specified goals, conversational analysis, or dashboard generation. Less attention has been given to how autonomous discovery agents can operate continuously over real-time streams, coordinate through production data infrastructure, and expose typed intermediate artifacts for observability, validation, and deployment.

We propose a shift from \emph{task-driven analytics} to \emph{discovery-driven analytics}, where analytical objectives are generated, validated, and refined by the system itself. Our approach decomposes the analytics lifecycle into specialized agents that transform data into hypotheses, executable analytics, validation reports, visualization specifications, and deployable applications. More broadly, we study the following operational model: \emph{analytics as continuous autonomous discovery over real-time data streams}.

This paper makes three contributions. First, we formulate autonomous insight discovery as a continuous agentic workload over streaming data. Second, we propose a contract-driven architecture in which agents exchange typed artifacts for hypotheses, analytic plans, generated code, validation reports, visualization specifications, and deployment manifests. Third, we discuss representative use cases in retail, finance, and public data, highlighting early lessons for observability, validation, and human oversight in agent-driven analytics systems.

\begin{figure*}[t]
    \centering
    \includegraphics[width=0.95\linewidth]{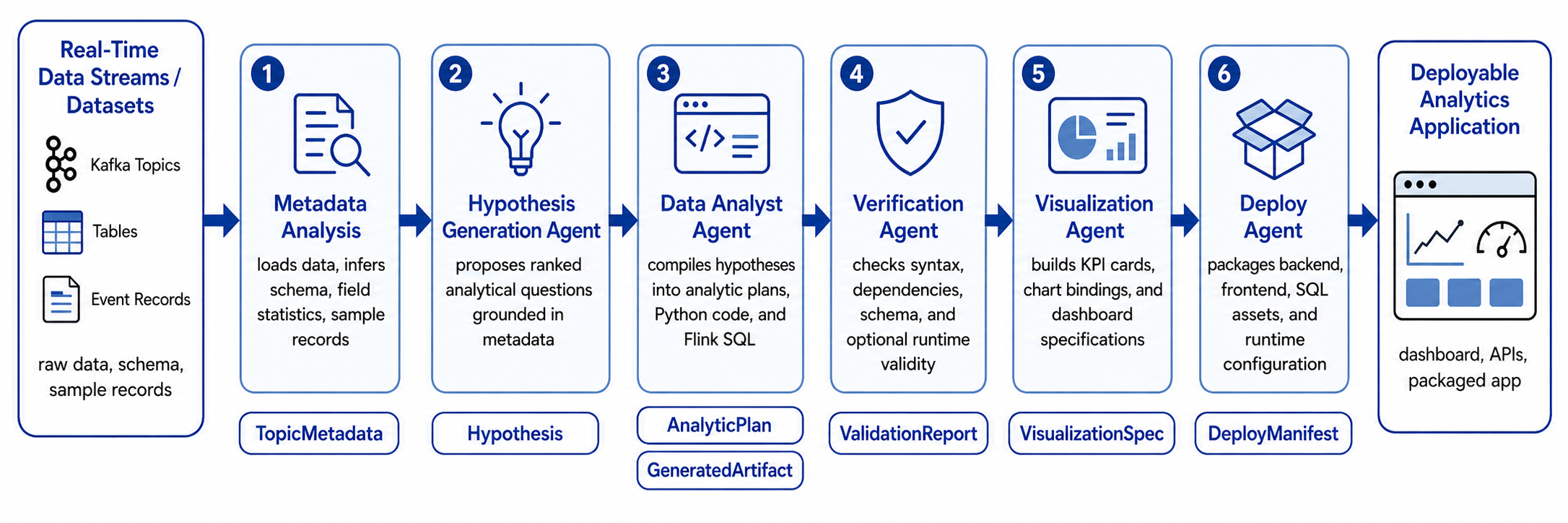}
    \Description{Architecture diagram showing real-time data streams and datasets flowing through metadata extraction, hypothesis generation, analytic planning, code generation, validation, visualization, and deployment stages.}
    \caption{Contract-driven discovery-agent pipeline for real-time analytics. Real-time data streams or datasets are first converted into structured metadata, then transformed by specialized agents into ranked hypotheses, analytic plans, executable Python and FlinkSQL~\cite{carbone2015apache} artifacts, validation reports, visualization specifications, and finally a deployable analytics application. Each stage produces typed intermediate artifacts that support lineage, observability, verification, and human oversight.}
    \label{fig:discovery-agents-architecture}
\end{figure*}

\section{Related Work}

LLM-powered data agents are emerging as a general abstraction for automating data management, preparation, analysis, visualization, and reporting. Recent surveys characterize these systems as tool-using agents that combine planning, retrieval, code generation, execution, and reflection across the data lifecycle~\cite{rahman2025datascienceagents,sun2025llmstatistics,zhu2025dataagents,fu2025autonomousdataagents}, while recent systems demonstrate their potential for heterogeneous data analysis, business intelligence, skill selection, external knowledge retrieval, and multi-agent reasoning~\cite{sun2025agenticdata,liu2025datasage,abaskohi2025agentada,jiang2025siriusbi}. Related work on automated visualization, insight management, and data-to-dashboard generation shows how LLMs can produce charts, visual explanations, dashboard specifications, and navigable insight spaces from data and user intent~\cite{zhang2025datatodashboard,weng2025insightlens,wu2024nl2vis,hoque2025nlgvis,chen2025coda}. In parallel, benchmarks and discovery-oriented systems increasingly frame insight generation as the target capability, evaluating agents on multi-step discovery, open-ended data analysis, business insight generation, heterogeneous data sources, and scientific hypothesis exploration~\cite{majumder2025discoverybench,gu2024blade,sahu2024insightbench,majumder2024datadrivendiscovery,liu2026datastorm,agarwal2025autodiscovery,mitchener2025kosmos}. Finally, work at the intersection of LLMs and databases highlights persistent challenges around trust, correctness, efficiency, and safe execution~\cite{li2024llmdatamanagement,kim2024trustworthy}. Our work builds on these directions but differs in focus: rather than targeting static datasets, user-specified goals, reports, or interactive dashboards, we propose a streaming, contract-driven architecture in which agents continuously generate hypotheses, compile analytics, validate artifacts, produce visualizations, and package the result into deployable applications.

\section{Architecture: Discovery Agents over Streaming Data}

The proposed system implements a multi-agent architecture for transforming real-time data streams into deployable analytics applications. At a high level, the system follows the pipeline shown in Figure~\ref{fig:discovery-agents-architecture}:

\begin{center}
\textit{Data $\rightarrow$ Metadata $\rightarrow$ Hypotheses $\rightarrow$ Analytics $\rightarrow$ Validation $\rightarrow$ Visualization $\rightarrow$ Application}
\end{center}

Unlike traditional data pipelines, where transformations are specified in advance, the computational workflow is constructed dynamically at runtime. Raw data is first converted into a semantic representation, \texttt{TopicMetadata}, capturing schema information, field statistics, sample records, and contextual metadata. This representation is then used to generate analytical hypotheses, which are compiled into executable artifacts, validated, visualized, and finally packaged into deployable applications.

A central design principle is the use of \emph{typed artifact contracts}. Each agent consumes and produces explicit intermediate representations such as \texttt{Hypothesis}, \texttt{AnalyticPlan}, \texttt{GeneratedArtifact}, \texttt{ValidationReport}, \texttt{VisualizationSpec}, and \texttt{DeployManifest}.\\
These contracts make the discovery process modular, inspectable, and reproducible.

The architecture supports both batch and streaming execution. In streaming deployments, Kafka acts as the communication backbone between agents, enabling asynchronous coordination and parallel exploration of hypotheses. Flink supports execution of generated SQL artifacts over continuous streams and provides operational visibility into the analytical pipeline. Together, these components allow the system to operate as a continuous discovery loop, where new data can trigger new hypotheses and updated analytics without manual query authoring.

\section{Agent Semantics and Artifact Contracts}

The system decomposes the analytics lifecycle into specialized agents, each responsible for a semantic transformation over typed artifacts. This design avoids monolithic LLM workflows and makes the discovery process observable at each stage.

\subsection{Hypothesis Generation Agent}

The Hypothesis Generation Agent initiates the discovery process by proposing analytical questions from data characteristics. It consumes \texttt{TopicMetadata}, including schema definitions, field statistics, sample records, and optional domain annotations.

The agent is grounded in the taxonomy of data-analysis questions proposed by Leek and Peng~\cite{leek2015question}, which distinguishes descriptive, exploratory, inferential, predictive, causal, and mechanistic questions. This taxonomy is useful because different question types imply different analytical assumptions and methods: descriptive questions summarize observed data, exploratory questions search for patterns, inferential questions ask whether patterns generalize, predictive questions estimate future or unseen outcomes, causal questions ask about interventions, and mechanistic questions seek to explain underlying processes.

Before generating hypotheses, the agent categorizes fields into analytical roles, such as numerical, categorical, temporal, or textual attributes. This grounding step constrains generation to questions that can be expressed using available data operations. For example, numerical fields may support descriptive summaries or anomaly detection, categorical fields may support exploratory group comparisons, and temporal fields may support predictive trend analysis.

The output is a set of structured \texttt{Hypothesis} objects:

\begin{verbatim}
{
  "category": "descriptive",
  "question": "...",
  "rationale": "...",
  "expected_insight": "...",
  "priority": 8
}
\end{verbatim}

Each hypothesis includes a category, rationale, expected insight, and priority score. This structure allows downstream agents to select appropriate analytical strategies and focus computational resources on higher-value questions.

\subsection{Data Analyst Agent}

The Data Analyst Agent translates hypotheses into executable analytics. Given a \texttt{Hypothesis} and the corresponding \texttt{TopicMetadata}, it produces an \texttt{AnalyticPlan} and one or more \texttt{GeneratedArtifact} objects.

A key design choice is dual artifact generation. Python artifacts provide flexible batch execution and support complex analytical logic, while FlinkSQL artifacts express equivalent computations for continuous stream processing. This allows the same analytical intent to be evaluated both offline and over live data streams.

The agent first constructs an analytic plan that identifies the required fields, the intended computation, the expected output schema, and the assumptions needed to answer the hypothesis. This planning step separates \emph{what} should be computed from \emph{how} it is implemented, making the generated code easier to validate, debug, and regenerate when failures occur.

Generated Python artifacts follow a strict runtime contract:

\begin{verbatim}
def analyze(data_records):
    return {"results": [...]}
\end{verbatim}

The output structure is standardized so that downstream validation, visualization, and deployment agents can consume results without relying on ad hoc parsing. Conceptually, the Data Analyst Agent acts as a compiler from analytical intent to executable programs.

This compiler-like role is important because different hypotheses require different analytical operators. Descriptive hypotheses may compile into aggregations and summaries, exploratory hypotheses into group comparisons or correlations, predictive hypotheses into trend or forecasting logic, and temporal hypotheses into windowed computations. By making these choices explicit in the \texttt{AnalyticPlan}, the system preserves a traceable link between the original question, the selected analytical method, and the generated executable artifact.

\subsection{Verification Agent}

The Verification Agent acts as a quality gate for generated artifacts. Since LLM-generated code may contain syntax errors, invalid assumptions, unsafe imports, or schema mismatches, validation is treated as a first-class stage rather than an optional post-processing step. This stage turns artifact correctness into an explicit, inspectable object, rather than leaving reliability as an implicit property of the generated code.

The agent performs syntax checks, dependency checks, schema validation, and, when sample data is available, runtime validation. The output is a structured \texttt{ValidationReport}:

\begin{verbatim}
{
  "status": "VALIDATED",
  "syntax_check": true,
  "schema_check": true,
  "runtime_check": false,
  "issues": []
}
\end{verbatim}

Validation issues are classified as errors, warnings, or informational messages. Errors prevent artifacts from progressing, while warnings allow execution with caveats. Rejected artifacts can trigger regeneration with validation feedback, enabling iterative refinement.

\subsection{Visualization Agent}

The Visualization Agent converts validated analytical outputs into dashboard specifications. Rather than requiring manual chart design, the agent inspects output schemas and, when possible, sample execution results to infer appropriate visual encodings. It maps the semantic shape of the result---for example categorical distributions, temporal trends, scalar summaries, or ranked lists---to dashboard elements such as bar charts, line charts, KPI cards, tables, and summary views.

The output is a \texttt{VisualizationSpec} describing charts, KPI cards, data bindings, and layout:

\begin{verbatim}
{
  "charts": [
    {
      "type": "bar",
      "x": "category",
      "y": "count"
    }
  ],
  "kpis": [...]
}
\end{verbatim}

This specification separates analytical computation from presentation. It also enables multiple frontends to render the same discovered insight while preserving a traceable connection to the originating hypothesis and generated artifact.

\subsection{Deploy Agent}

The Deploy Agent materializes the validated analytics and visualization specifications into a runnable application. It packages generated code, dashboard configuration, backend services, frontend components, and deployment metadata. It also preserves the connection between runtime assets and upstream artifacts, so that deployed dashboards can be traced back to the hypotheses, code, validation reports, and visualization specifications that produced them.

The output is a \texttt{DeployManifest} that records generated files, configuration parameters, artifact lineage, and deployment instructions. This manifest serves as the final contract between the discovery pipeline and the runtime environment.
In this way, the system does not merely generate insights; it packages them into inspectable applications that can be executed, reviewed, and extended by users.

\subsection{Contract-Driven Coordination}

Typed artifact contracts provide the coordination layer across agents. Each transformation is explicit:

\begin{center}
\small
\texttt{TopicMetadata}
$\rightarrow$
\texttt{Hypothesis}
$\rightarrow$
\texttt{AnalyticPlan}
$\rightarrow$
\texttt{GeneratedArtifact}
$\rightarrow$
\texttt{ValidationReport}
$\rightarrow$
\texttt{VisualizationSpec}
$\rightarrow$
\texttt{DeployManifest}
\end{center}

This structure enables lineage tracking across the discovery process. Hypotheses reference their source topics, generated artifacts reference the hypotheses they implement, validation reports reference the artifacts they evaluate, and deployment manifests reference all upstream objects.

In streaming deployments, these artifacts can be exchanged through Kafka topics, allowing multiple hypotheses to be explored in parallel. This creates a natural execution model for agentic analytics: agents operate asynchronously, artifacts provide the shared state, and validation reports control progression through the pipeline.
The same contract structure also creates intervention points for human oversight. Users can review hypotheses before execution, inspect generated code before deployment, examine validation reports, or adjust visualization specifications. This balances automation with control and makes the system suitable for production-oriented analytics settings.

\section{Use Cases and Early Lessons}

Given the exploratory nature of the system, we evaluate it through representative use cases rather than a formal benchmark. The goal is to characterize the types of discovery workflows enabled by the architecture and identify early design lessons.

\subsection{Retail Analytics}

In retail analytics, the system processes transaction streams containing product, category, and customer interaction data. The Hypothesis Generation Agent identifies candidate patterns such as category dominance, temporal purchasing trends, basket composition, and anomalies in sales distribution.
The system then generates aggregation queries, time-series analyses, and dashboards highlighting indicators such as top-selling categories, changes in product mix, or distribution shifts. This illustrates how proactive hypothesis generation can reduce the manual effort required to design exploratory dashboards.

\subsection{Financial Monitoring}

In financial monitoring, the system operates over transaction or market-event streams and proposes hypotheses related to volume spikes, unusual activity distributions, deviations from expected behavior, or temporal concentration of events.
Generated analytics can surface candidate anomalies and summarize them through interpretable dashboards. Continuous hypothesis generation is particularly useful in this setting because relevant patterns may evolve rapidly and may not be known in advance.

\subsection{Governance and Public Data}

In public data scenarios, such as NYC Open Data, the system operates over heterogeneous datasets related to public services, infrastructure, complaints, or demographic indicators. It can propose hypotheses about temporal trends, geographic differences, or relationships between service requests and neighborhoods.
The resulting dashboards can help analysts and policymakers explore datasets without first defining a fixed set of queries. This use case highlights the value of autonomous discovery in broad, heterogeneous data environments where the relevant analytical questions may not be obvious upfront.

\subsection{Cross-Domain Lessons}

Across these scenarios, three lessons emerge. First, hypothesis generation must be grounded in metadata and field statistics to avoid irrelevant or non-executable questions. Second, validation must be treated as a first-class stage, since generated code and queries may fail syntactically, semantically, or operationally. Third, observability over intermediate artifacts is essential: users and operators need to inspect not only final dashboards, but also the hypotheses, plans, code, and validation reports that produced them.
These lessons suggest that agentic analytics systems require more than LLM prompting. They require system-level support for contracts, lineage, verification, and human oversight.

\section{Position and Open Challenges}

We argue that proactive insight discovery should become a first-class workload for agent-first data systems. The proposed architecture shifts analytics from a query-driven process, where goals are specified externally, to a discovery-driven process, where agents generate and prioritize analytical objectives directly from data.
This elevates autonomy from execution to \emph{problem formulation}. Instead of acting only as a query processor, the system operates as a continuous discovery engine. Analytical objectives emerge from the interaction between data, metadata, validation feedback, and specialized agents.
The contract-driven design is a key enabler of this transition. Typed artifacts make LLM-based generation more controllable by exposing intermediate representations that can be validated, logged, inspected, and replayed. This is especially important in streaming environments, where agent actions may be triggered continuously and must remain auditable.

Several open challenges remain. First, hypothesis quality is difficult to evaluate: a valid hypothesis may still be uninteresting, redundant, or misleading. Second, generated analytics must be checked not only for syntax, but also for statistical validity and operational safety. Third, continuous discovery may produce many candidate insights, requiring ranking, deduplication, and human-in-the-loop review. Finally, deployment raises governance questions around provenance, access control, and accountability for system-generated conclusions.
These challenges suggest a broader research agenda for agentic data systems: building infrastructure that supports not only faster query answering, but also reliable, observable, and steerable autonomous exploration.

\section{Conclusion}

We presented a multi-agent architecture for autonomous discovery over real-time data streams. By combining Kafka-based coordination, Flink-based stream processing, and LLM-powered agents with typed artifact contracts, the system transforms data into hypotheses, analytics, validation reports, visualizations, and deployable applications.
The central contribution is a shift from automating analysis to automating discovery. Rather than accelerating only user-specified workflows, the proposed architecture treats insight generation itself as a continuous, agentic process embedded within the data infrastructure.

\appendix

\section{Generated Artifacts: NYC Parks Events Demo}
\label{app:nyc-parks-demo}

This appendix summarizes the artifacts generated by the discovery-agent pipeline for the NYC Parks Events dataset~\footnote{\url{https://data.cityofnewyork.us/City-Government/NYC-Parks-Events-Listing-Event-Listing/fudw-fgrp/about_data}}. The goal is not to document every generated file exhaustively, but to show how the proposed architecture materializes the discovery loop into inspectable, executable, and deployable artifacts.

\subsection{Demo Summary}

The NYC Parks Events Listing dataset contains public event records across New York City parks. Each record includes temporal, geographic, textual, categorical, and cost-related attributes. The pipeline processed the dataset through the full discovery loop:

\begin{center}
\small
\texttt{Metadata}
$\rightarrow$
\texttt{Hypotheses}
$\rightarrow$
\texttt{Analytics}
$\rightarrow$
\texttt{Validation}
$\rightarrow$
\texttt{Visualization}
$\rightarrow$
\texttt{Deployment}
\end{center}

Table~\ref{tab:appendix-demo-summary} summarizes the input data and generated outputs.

\begin{table*}[t]
\centering
\caption{Summary of the NYC Parks Events discovery demo.}
\label{tab:appendix-demo-summary}
\small
\begin{tabular}{p{0.18\linewidth}p{0.74\linewidth}}
\toprule
\textbf{Component} & \textbf{Description} \\
\midrule
Dataset & NYC Parks Events Listing dataset, containing scheduled public events across parks in New York City. \\
\addlinespace
Representative fields & \texttt{date}, \texttt{start\_time}, \texttt{end\_time}, \texttt{location}, \texttt{borough}, \texttt{park\_name}, \texttt{title}, \texttt{description}, \texttt{event\_type}, \texttt{cost\_free}, \texttt{must\_see}. \\
\addlinespace
Generated hypotheses & Analytical questions covering descriptive, exploratory, inferential, and predictive categories. \\
\addlinespace
Generated analytics & Python artifacts for batch analysis and FlinkSQL artifacts for stream-oriented execution. \\
\addlinespace
Validation outputs & Structured validation reports covering syntax, dependency, schema, and runtime checks. \\
\addlinespace
Generated interface & Dashboard with KPI cards, hypothesis tabs, and interactive visualizations. \\
\addlinespace
Deployment output & Containerized application with backend API, frontend dashboard, source data, generated artifacts, and deployment manifest. \\
\bottomrule
\end{tabular}
\end{table*}

\subsection{End-to-End Lineage Example}

To illustrate the full discovery pipeline, we trace a single high-priority hypothesis through all transformation stages. This example shows how typed artifact contracts preserve lineage from an analytical question to the deployed application.

\subsubsection{Stage 1: Hypothesis Generation}

The Hypothesis Generation Agent produced the following hypothesis from dataset metadata:

\begin{verbatim}
{
  "id": "hyp_20260518_133512_090881",
  "created_at": "2026-05-18T13:35:12.090893",
  "source_topic_id": "topic_20260518_133450_224571",
  "category": "descriptive",
  "question": "What is the temporal distribution of NYC
    Parks events across years, months, and days of
    the week, and are there clear seasonal patterns
    in event scheduling?",
  "rationale": "With 74,880 events spanning multiple
    years, understanding when events are concentrated
    reveals how NYC Parks allocates programming
    resources across seasons.",
  "expected_insight": "We expect to find strong
    seasonality with peaks in spring/summer months
    and troughs in winter, with weekends having
    significantly more events than weekdays.",
  "priority": 9
}
\end{verbatim}

The hypothesis references the source topic and includes a priority score used to rank candidate analyses.

\subsubsection{Stage 2: Analytic Plan Generation}

The Data Analyst Agent translated the hypothesis into an executable plan:

\begin{verbatim}
{
  "id": "plan_20260518_133551_981283",
  "hypothesis_id": "hyp_20260518_133512_090881",
  "approach": "Parse the date field from each event
    record to extract year, month, and day of week.
    Aggregate event counts by these temporal
    dimensions to reveal distribution patterns.",
  "steps": [
    "Load and validate data",
    "Parse date strings into datetime objects",
    "Extract year, month, and day_of_week",
    "Aggregate event counts by temporal dimensions",
    "Compute seasonal distribution",
    "Calculate weekday versus weekend statistics",
    "Return structured results"
  ],
  "output_schema": {
    "results": "dict with keys: [summary,
      yearly_distribution, monthly_distribution,
      day_of_week_distribution, seasonal_distribution,
      weekday_weekend_comparison]"
  }
}
\end{verbatim}

The plan separates analytical intent from implementation. This makes the generated code easier to validate, debug, regenerate, and explain.

\subsubsection{Stage 3: Python Artifact Generation}

The agent generated a Python artifact implementing the analytic plan:

\begin{verbatim}
{
  "id": "artifact_py_20260518_133551_981378",
  "hypothesis_id": "hyp_20260518_133512_090881",
  "analytic_plan_id": "plan_20260518_133551_981283",
  "artifact_type": "python_code",
  "language": "python",
  "dependencies": ["pandas", "numpy"]
}
\end{verbatim}

The generated function follows the standard runtime contract: it receives records as input and returns a structured result object.

\begin{lstlisting}[
    language=Python,
    caption={Generated Python artifact for temporal distribution analysis.},
    label={lst:appendix-python-artifact},
    basicstyle=\footnotesize\ttfamily,
    breaklines=true
]
def analyze_temporal_distribution(
    data_records: List[Dict[str, Any]]
) -> Dict[str, Any]:
    """Analyze temporal distribution of NYC Parks events."""
    df = pd.DataFrame(data_records)

    # Parse dates and keep valid records.
    df["parsed_date"] = df["date"].apply(parse_date)
    df_valid = df[df["parsed_date"].notna()].copy()

    # Derive temporal features.
    df_valid["year"] = df_valid["parsed_date"].dt.year
    df_valid["month"] = df_valid["parsed_date"].dt.month
    df_valid["day_of_week"] = df_valid["parsed_date"].dt.dayofweek
    df_valid["season"] = df_valid["month"].apply(get_season)

    # Compute distributions.
    yearly_distribution = (
        df_valid.groupby("year")
        .size()
        .reset_index(name="event_count")
    )

    monthly_distribution = (
        df_valid.groupby("month")
        .size()
        .reset_index(name="event_count")
    )

    seasonal_distribution = (
        df_valid.groupby("season")
        .size()
        .reset_index(name="event_count")
    )

    return {
        "results": {
            "summary": {...},
            "yearly_distribution": yearly_distribution,
            "monthly_distribution": monthly_distribution,
            "seasonal_distribution": seasonal_distribution
        }
    }
\end{lstlisting}

The key property is that the artifact does not return free-form text. It returns a structured object that can be validated, cached, visualized, and served by the generated application.

\subsubsection{Stage 4: FlinkSQL Artifact Generation}

The agent also generated a FlinkSQL artifact for stream-oriented execution of the same analytical intent:

\begin{verbatim}
{
  "id": "artifact_sql_20260518_133551_981443",
  "hypothesis_id": "hyp_20260518_133512_090881",
  "analytic_plan_id": "plan_20260518_133551_981283",
  "artifact_type": "flink_sql"
}
\end{verbatim}

\begin{lstlisting}[
    language=SQL,
    caption={Generated FlinkSQL artifact for temporal aggregation.},
    label={lst:appendix-flink-artifact},
    basicstyle=\footnotesize\ttfamily,
    breaklines=true
]
SELECT
    EXTRACT(YEAR FROM TO_DATE(`date`, 'MM/dd/yyyy'))
        AS event_year,
    EXTRACT(MONTH FROM TO_DATE(`date`, 'MM/dd/yyyy'))
        AS event_month,
    CASE
        WHEN EXTRACT(MONTH FROM TO_DATE(`date`,
            'MM/dd/yyyy')) IN (12, 1, 2) THEN 'Winter'
        WHEN EXTRACT(MONTH FROM TO_DATE(`date`,
            'MM/dd/yyyy')) IN (3, 4, 5) THEN 'Spring'
        WHEN EXTRACT(MONTH FROM TO_DATE(`date`,
            'MM/dd/yyyy')) IN (6, 7, 8) THEN 'Summer'
        ELSE 'Fall'
    END AS season,
    COUNT(*) AS event_count
FROM `NYC_Parks_Events_Listing_-_Event_Listing`
WHERE `date` IS NOT NULL AND `date` <> ''
GROUP BY event_year, event_month, season;
\end{lstlisting}

This dual-artifact pattern is central to the architecture: Python supports flexible batch execution, while FlinkSQL provides a path toward continuous analytics over streams.

\subsubsection{Stage 5: Validation}

The Verification Agent validated the generated artifact before it was passed to visualization and deployment:

\begin{verbatim}
{
  "id": "validation_20260518_133936_698650",
  "artifact_id": "artifact_py_20260518_133551_981378",
  "status": "validated",
  "syntax_check": true,
  "import_check": true,
  "schema_check": true,
  "runtime_check": false,
  "issues": [],
  "feedback": "Validation passed successfully"
}
\end{verbatim}

The validation report confirms that the generated artifact passes syntax, dependency, and schema checks. Failed validations can trigger regeneration with structured feedback, making validation part of the agentic workflow rather than a manual post-processing step.

\subsubsection{Stage 6: Visualization Specification}

The Visualization Agent mapped the validated output to dashboard components. Each component references a specific artifact and output path, preserving traceability from visual element to computation.

\begin{verbatim}
{
  "id": "viz_20260518_133953_921433",
  "hypothesis_ids": ["hyp_20260518_133512_090881", ...],
  "artifact_ids": [
    "artifact_py_20260518_133551_981378",
    "artifact_sql_20260518_133551_981443",
    ...
  ],
  "dashboard": {
    "kpi_cards": [
      {
        "id": "kpi_1",
        "title": "Total Events Analyzed",
        "value_field": "artifacts.artifact_py_20260518_
          133551_981378.summary.total_events_analyzed"
      }
    ],
    "charts": [
      {
        "id": "chart_1",
        "title": "Yearly Event Distribution",
        "chart_type": "bar",
        "data_source": "artifacts.artifact_py_20260518_
          133551_981378.yearly_distribution",
        "x_field": "year",
        "y_field": "event_count"
      },
      {
        "id": "chart_2",
        "title": "Monthly Event Distribution",
        "chart_type": "line",
        "data_source": "artifacts.artifact_py_20260518_
          133551_981378.monthly_distribution"
      }
    ]
  }
}
\end{verbatim}

\begin{figure*}[t]
    \centering
    \fbox{\includegraphics[width=0.95\linewidth]{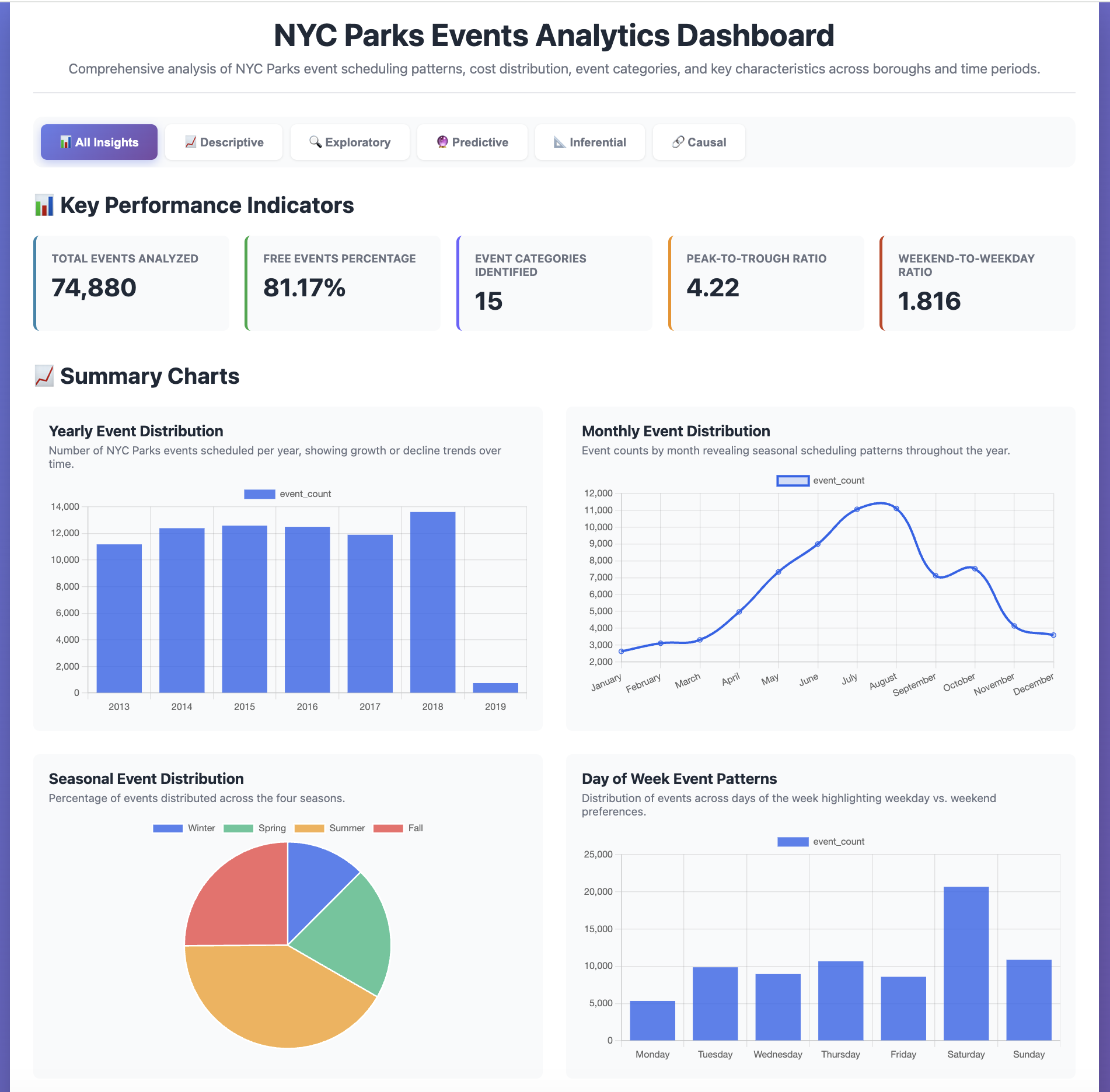}}
    \caption{Generated dashboard for NYC Parks Events analytics. The interface includes KPI cards, hypothesis tabs, and automatically selected visualizations derived from validated analytical outputs.}
    \label{fig:dashboard-screenshot}
\end{figure*}

Figure~\ref{fig:dashboard-screenshot} shows the generated interface.

\subsubsection{Stage 7: Deployment Manifest}

Finally, the Deploy Agent packaged the generated artifacts into a deployable application:

\begin{verbatim}
{
  "id": "deploy_20260518_133954_020895",
  "source_topic_id": "topic_20260518_133450_224571",
  "hypothesis_ids": ["hyp_20260518_133512_090881", ...],
  "artifact_ids": [
    "artifact_py_20260518_133551_981378",
    "artifact_sql_20260518_133551_981443",
    ...
  ],
  "visualization_spec_id": "viz_20260518_133953_921433",
  "app_name": "nyc-park-events-demo",
  "generated_files": [
    {
      "path": "backend/app/analytics/
        artifact_py_20260518_133551_981378.py",
      "type": "python",
      "description": "Python analysis for temporal
        distribution hypothesis"
    },
    {
      "path": "backend/app/sql/
        artifact_sql_20260518_133551_981443.sql",
      "type": "sql",
      "description": "FlinkSQL for temporal
        distribution hypothesis"
    }
  ]
}
\end{verbatim}

The deployment manifest preserves the complete lineage chain. Any deployed dashboard element can be traced back through visualization specifications, validation reports, generated code, analytic plans, the originating hypothesis, and the source data.

\subsection{Additional Generated Hypotheses}

Table~\ref{tab:appendix-hypotheses} summarizes additional high-priority hypotheses generated for the demo. These hypotheses follow the same transformation pipeline illustrated above.

\begin{table*}[t]
\centering
\caption{Additional representative hypotheses generated for the NYC Parks Events dataset.}
\label{tab:appendix-hypotheses}
\small
\begin{tabular}{p{0.07\linewidth}p{0.18\linewidth}p{0.65\linewidth}}
\toprule
\textbf{ID} & \textbf{Category} & \textbf{Question} \\
\midrule
H2 & Exploratory & What proportion of events are free versus paid, and how does this ratio vary by event type, location, and time of year? \\
\addlinespace
H3 & Descriptive & What are the most common event categories, and how are they distributed across parks and boroughs? \\
\addlinespace
H4 & Inferential & Is there a relationship between an event being marked as \texttt{must\_see} and characteristics such as cost, location, time of day, or event type? \\
\addlinespace
H5 & Predictive & Can event duration be predicted from event type, location, day of week, and whether the event is free? \\
\bottomrule
\end{tabular}
\end{table*}

\subsection{Generated Application Structure}

The Deploy Agent packages the generated artifacts into a runnable application. The resulting directory contains the backend service, frontend dashboard, source data, generated analytics, SQL artifacts, validation reports, and deployment metadata.

\begin{verbatim}
nyc-park-events-demo/
|-- README.md
|-- docker-compose.yml
|-- deploy_manifest.json
|-- VERIFICATION_REPORT.txt
|-- backend/
|   |-- Dockerfile
|   |-- requirements.txt
|   `-- app/
|       |-- main.py
|       |-- analytics/
|       |   `-- artifact_py_*.py
|       `-- sql/
|           |-- artifact_sql_*.sql
|           `-- manifest.json
|-- frontend/
|   |-- Dockerfile
|   |-- nginx.conf
|   |-- public/
|   |   `-- index.html
|   `-- src/
|       |-- dashboard.js
|       `-- styles.css
`-- data/
    `-- full_dataset.csv
\end{verbatim}

The backend exposes REST endpoints for health checks, dataset summaries, KPI values, chart data, data previews, and hypothesis metadata. The frontend consumes these endpoints and renders the generated dashboard.

\subsection{Execution and Discovered Insights}

The application can be launched with:

\begin{verbatim}
docker-compose up --build
\end{verbatim}

This starts the backend and frontend services, loads the dataset, executes the generated analytical functions, caches results, and serves them through the dashboard.

The generated analysis surfaced several representative insights:

\begin{itemize}
    \item \textbf{Temporal patterns}: Events are concentrated in warmer months, with stronger activity during summer.
    \item \textbf{Cost distribution}: Most events are free, with variation across event types and locations.
    \item \textbf{Event categories}: Frequent categories include fitness, cultural programming, nature education, and family activities.
    \item \textbf{Weekend preference}: Events are more concentrated on weekends than weekdays.
    \item \textbf{Borough distribution}: A large share of events is concentrated in Manhattan and Brooklyn.
\end{itemize}

These results illustrate the intended behavior of the system: the user does not manually write the analytical queries. Instead, the system proposes hypotheses, generates analytics, validates outputs, and packages the results into an application.

\subsection{Lessons from the Demo}

The NYC Parks Events demo highlights five practical lessons.

\begin{enumerate}
    \item \textbf{Typed contracts improve reliability.} Explicit intermediate artifacts make the discovery process easier to inspect, validate, and replay.
    \item \textbf{Hypothesis generation must be grounded.} Metadata and field statistics are necessary to avoid irrelevant or non-executable questions.
    \item \textbf{Validation is essential.} Generated code and SQL must be checked before they are used in downstream visualizations or deployed applications.
    \item \textbf{Dual artifacts support multiple execution modes.} Python artifacts support batch analytics, while FlinkSQL artifacts provide a path toward streaming execution.
    \item \textbf{Lineage enables oversight.} Deployment manifests preserve the connection between hypotheses, code, validation reports, visualizations, and application files.
\end{enumerate}

Overall, the demo shows that the proposed architecture does more than generate isolated insights. It produces a complete, inspectable chain from dataset metadata to a deployable analytics application.

\bibliographystyle{ACM-Reference-Format}
\bibliography{references}

\end{document}